# Bootstrapping Transliteration with Constrained Discovery for Low-Resource Languages

**Shyam Upadhyay**
University of Pennsylvania
Philadelphia, PA
shyamupa@seas.upenn.edu

**Jordan Kodner**
University of Pennsylvania
Philadelphia, PA
jkodner@seas.upenn.edu

**Dan Roth**
University of Pennsylvania
Philadelphia, PA
danroth@seas.upenn.edu

## Abstract

*Generating* the English transliteration of a name written in a foreign script is an important and challenging step in multilingual knowledge acquisition and information extraction. Existing approaches to transliteration generation require a large (>5000) number of training examples. This difficulty contrasts with transliteration *discovery*, a somewhat easier task that involves picking a plausible transliteration from a given list. In this work, we present a bootstrapping algorithm that uses constrained discovery to improve generation, and can be used with as few as 500 training examples, which we show can be sourced from annotators in a matter of hours. This opens the task to languages for which large number of training examples are unavailable. We evaluate transliteration generation performance itself, as well the improvement it brings to cross-lingual candidate generation for entity linking, a typical downstream task. We present a comprehensive evaluation of our approach on nine languages, each written in a unique script.[1]

## 1 Introduction

*Transliteration* is the process of transducing names from one writing system to another (e.g., ओबामा in Devanagari to *Obama* in Latin script) while preserving their pronunciation (Knight and Graehl, 1998; Karimi et al., 2011). In particular, *back-transliteration* from foreign languages to English has applications in multilingual knowledge acquisition tasks including named entity recognition (Darwish, 2013) and information retrieval (Virga and Khudanpur, 2003). Two tasks feature prominently in the transliteration literature: *generation* (Knight and Graehl, 1998) which involves producing an appropriate transliteration for a given word in an open-ended way, and *discovery* (Sproat et al., 2006; Klementiev and Roth, 2008) which involves selecting an appropriate transliteration for a word from a list of candidates. This work develops transliteration generation approaches for low-resource languages.

Existing transliteration generation models require supervision in the form of source-target name pairs (≈5-10k), which are often collected from names in Wikipedia inter-language links (Irvine et al., 2010). However, most languages that use non-Latin scripts are under-represented in terms of such resources. Table 1 illustrates this issue, and the extra coverage one can achieve by extending to low-resource languages. A model that requires 50k name pairs as supervision can only support 6 languages, while one that just needs 500 could support 56. For a model to be widely applicable, it must function in low-resource settings.

| # Name Pairs in Wikipedia | Languages | Scripts |
|---|---|---|
| > 50,000 | 6 | 5 |
| > 10,000 | 18 | 14 |
| > 5,000  *Previous Work* | 24 | 15 |
| > 1,000 | 45 | 22 |
| > 500  *Our Approach* | 56 | 23 |
| > 0 | 93 | 30 |

Table 1: Cumulative number of person name pairs in Wikipedia inter-language links. While previous approaches for transliteration generation were applicable to only 24 languages (spanning 15 scripts), our approach is applicable to 56 languages (23 scripts). When counting scripts we exclude variants (e.g., all Cyrillic scripts and variants count as one).

We propose a new bootstrapping algorithm that uses a weak generation model to guide discovery of good transliterations, which in turn aids future bootstrapping iterations.[2] By carefully controlling the interaction of discovery and the generation model via constrained inference, we show how to bootstrap a generation model using a dictionary of names in English, a list of words in the foreign script, and little initial supervision (≈500 name pairs). To the best of our knowledge, ours is the first work to accomplish transliteration generation in such a low-resource setting.

---

[1]code at github.com/shyamupa/hma-translit.

[2]All generative approaches are also capable of discovery, by using the posterior P(y | x) to select the most likely candidate transliteration, while the opposite is not true.

We demonstrate the practicality of our approach in truly low-resource scenarios and downstream applications through two case studies. First, in §8.1 we show that one can obtain the initial supervision from a *single* human annotator within a few hours for two languages – Armenian and Punjabi. This is a realistic scenario where language access is limited to a single native informant. Second, in §8.2 we show that our approach benefits a typical downstream application, namely candidate generation for cross-lingual entity linking, by improving recall on two low-resource languages – Tigrinya and Macedonian. We also present an analysis (§7) of the inherent challenges of transliteration, and the trade-off between native (i.e., source) and foreign (i.e., target) vocabulary.

## 2 Related Work

We briefly review the limitations of existing generation and discovery approaches, and provide an overview of how our work addresses them.

**Transliteration Generation** (Haizhou et al., 2004; Jiampojamarn et al., 2009; Ravi and Knight, 2009; Jiampojamarn et al., 2010; Finch et al., 2015, *inter alia*) requires generous amount of name pairs (≈5-10k) in order to learn to map words in the source script to the target script. While some approaches (Irvine et al., 2010; Tsai and Roth, 2018) use Wikipedia inter-language links to identify name pairs for supervision, a truly low-resource language (like Tigrinya) is likely to have limited Wikipedia presence as well.

**Transliteration Discovery** (Sproat et al., 2006; Chang et al., 2009) is considerably easier than generation, owing to the smaller search space. However, discovery often uses features derived from resources that are unavailable for low-resource languages, like comparable corpora (Sproat et al., 2006; Klementiev and Roth, 2008).

A key limitation of discovery is the assumption that the correct transliteration(s) is in the list of candidates $\mathcal{N}$. Since discovery models always pick *something* from $\mathcal{N}$, they can produce false positives, if no correct transliteration is present in $\mathcal{N}$.

To overcome this, it is prudent to develop generation models which can handle input for which the transliteration does not belong in $\mathcal{N}$.

**Our Work** We show that a weak generation model can be iteratively improved using constrained discovery. In particular, our work uses a weak generation model to discover new training pairs, using constraints to drive the bootstrapping. Our generation model is inspired by the success of sequence to sequence generation models (Sutskever et al., 2014; Bahdanau et al., 2015) for string transduction tasks like inflection and derivation generation (Faruqui et al., 2016; Cotterell et al., 2017; Aharoni and Goldberg, 2017; Makarov et al., 2017). Our bootstrapping framework can be viewed as an instance of constraint driven learning (Chang et al., 2007, 2012).

## 3 Transliteration Generation with Hard Monotonic Attention - Seq2Seq(HMA)

We view generation as a string transduction task and use a sequence to sequence (Seq2Seq) generation model that uses *hard monotonic attention* (Aharoni and Goldberg, 2017), henceforth referred to as Seq2Seq(HMA). During generation, Seq2Seq(HMA) directly models the *monotonic* source-to-target sequence alignments, using a pointer that attends to a *single* input character at a time. Monotonic attention is a natural fit for transliteration because even though the number of characters needed to represent a sound in the source and target language vary, the sequence of sounds is presented in the same order.[3] We review Seq2Seq(HMA) below, and describe how it can be applied to transliteration generation.

**Encoding Input Word** Let $\Sigma_f$ be the source alphabet and $\Sigma_e$ be the English alphabet. Let $x = (x_1, x_2, \cdots, x_n)$ denote an input word where each character $x_i \in \Sigma_f$. The characters are first encoded using a embedding matrix $\mathbf{W} \in \mathbb{R}^{|\Sigma_f| \times d}$ to get character embeddings $\boldsymbol{x}_1, \boldsymbol{x}_2, \cdots, \boldsymbol{x}_n$ where each $\boldsymbol{x}_i \in \mathbb{R}^d$. These embeddings are fed into a bidirectional RNN encoder to generate encoded vectors $\boldsymbol{h}_1, \boldsymbol{h}_2, \cdots, \boldsymbol{h}_n$ where each $\boldsymbol{h}_i \in \mathbb{R}^{2k}$, and $k$ is the size of output vector of the forward (and backward) encoder. The encoded vectors $\boldsymbol{h}_1, \boldsymbol{h}_2, \cdots, \boldsymbol{h}_n$ are then fed into the decoder.

---
[3]Many Indic scripts, that sometimes write vowels before the consonants they are pronounced after, seem to violate this claim, but Unicode representations of these scripts actually preserve the consonant-vowel order.

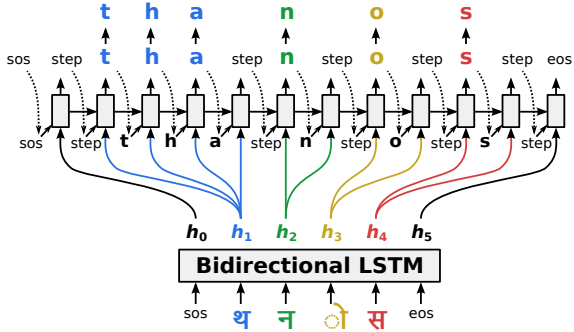

Figure 1: Transliteration using Seq2Seq transduction with Hard Monotonic Attention, or Seq2Seq(HMA). The figure shows how decoding proceeds for transliterating "थनोस" to "thanos". During decoding, the model attends to a source character (e.g.,थ shown in blue) and outputs target characters (t, h, a) until a $step$ action is generated, which moves the attention position forward by one character (to न), and so on.

**Monotonic Decoding with Hard Attention** Figure 1 illustrates the decoding process. The decoder RNN generates a sequence of actions $\{s_1, s_2, \cdots\}$, such that each $s_i \in \Sigma_e \cup \{step\}$. The $step$ action controls an attention position $a$, attending on input character $x_a$, with encoded vector $\boldsymbol{h}_a$. Each action $s_i$ is embedded into $\boldsymbol{s}_i \in \mathbb{R}^d$ using a output embedding matrix $\mathbf{A} \in \mathbb{R}^{(|\Sigma_e|+1) \times d}$. At any time during decoding, the decoder uses its last hidden state, the embedding of the previous action $\boldsymbol{s}_i$ and the encoded vector $\boldsymbol{h}_a$ of the current attended position to generate the next action $s_{i+1}$. If the generated action is $step$, the decoder increments the attention position by one. This ensures that the decoding is monotonic, as the attention position can only move forward or stay at the same position during generation. We use Inference$(G, x)$ to refer to the above decoding process for a trained generation model $G$ and input word $x$.

**Training** requires the oracle action sequence $\{s_i\}$ for input $x_{1:n}$ that generates the correct transliteration $y_{1:m}$. The oracle sequence is generated using the train name pairs and Algorithm 1 in Aharoni and Goldberg (2017), with the character-level alignment between $x_{1:n}$ and $y_{1:m}$ being generated using the algorithm in Cotterell et al. (2016).

**Inference Strategies** We describe an unconstrained and a constrained inference strategy to select the best transliteration $\hat{y}$ from a beam $\{y_i\}_{i=1}^k$ of transliteration hypotheses, sorted in descending order by likelihood. The constrained strategy use a name dictionary $\mathcal{N}$, to guide the inference. These strategies are applicable to any generation model.

- **Unconstrained (U)** selects the most likely item $y_1$ in the beam as $\hat{y}$.
- **Dictionary-Constrained (DC)** selects the highest scoring hypothesis that is present in $\mathcal{N}$, and defaults to $y_1$ if none are in $\mathcal{N}$.

It is tempting to disallow the model from generating hypotheses which are not in the dictionary $\mathcal{N}$. However, dictionaries are always incomplete, and restricting the search to generate from $\mathcal{N}$ inevitably leads to incorrect predictions if the correct transliteration is not in $\mathcal{N}$. This is essentially the same as the problem inherent to discovery models.

**Other Strategies in Previous Work** A related constrained inference strategy was proposed by Lin et al. (2016), who use a entity linking system (Wang et al., 2015) to correct and re-rank hypotheses, using any available context to aid hypothesis correction. Our constrained inference strategy is much simpler, requiring only a name dictionary $\mathcal{N}$. We experimentally show that our approach outperforms that of Lin et al. (2016).

## 4 Low-Resource Bootstrapping

Low-resource languages will have a limited number of name pairs for training a generation model. To learn a good generation model in this setting, we propose a new bootstrapping algorithm, that uses *constrained discovery* to mine name pairs to re-train the generation model. Our algorithm requires a small ($\approx$500) seed list of name pairs $\mathcal{S}$ for supervision, a dictionary $\mathcal{N}$ containing names in English, and a list of words $\mathcal{V}_f$ in the foreign script.

Below we describe our algorithm and the constraints used to guide discovery of new name pairs.

### 4.1 The Bootstrapping Algorithm

Algorithm 1 shows the pseudo-code of the bootstrapping procedure. We initialize a weak generation model $G_0$ using a seed list of name pairs $\mathcal{S}$ (line 1). At iteration $t$, the current generation model $G_t$ produces the top-$k$ transliteration hypotheses $\{y_i\}_{i=1}^k$ for each word $x \in \mathcal{V}_f$ (line 5). A source word and hypothesis pair $(x, y_i)$, is added to the set of mined name pairs $\mathcal{B}$ if they satisfy a set of discovery constraints (described below) (line 8). A new generation model $G_{t+1}$ is trained for the next iteration using the union of the seed list $\mathcal{S}$ and the mined name pairs $\mathcal{B}$ (line 12). $\mathcal{B}$ is purged after every iteration (line 3) to prevent $G_{t+1}$ from being influenced by possibly incorrect name pairs mined in

**Algorithm 1** Bootstrapping a Transliteration Generation Model via Constrained Discovery

**Input:**
    English name dictionary $\mathcal{N}$; Seed training pairs $\mathcal{S}$;
    Vocabulary in the target language $\mathcal{V}_f$.
**Hyper-parameters:**
    initial minimum length threshold $L_0^{min}$;
    minimum likelihood threshold $\delta^{min}$;
    length ratio tolerance $\epsilon$.
**Output:** Generation model $G_T$

1:  $G_0 = \texttt{train}(\mathcal{S})$   ▷ init. generation model.
2:  **while** not converged **do**
3:     $\mathcal{B} = \emptyset$   ▷ purge mined set.
4:     **for** $x$ in $\mathcal{V}_f$ **do**
5:         $\{y_i\}_{i=1}^k = \text{argtop}_k\, \texttt{Inference}(G_t, x)$
6:         **for** $y_i$ in $\{y_i\}_{i=1}^k$ **do**
7:             **if** $(x, y_i)$ satisfies constraints in §4.2 **then**
8:                 $\mathcal{B} = \mathcal{B} \cup \{(x, y_i)\}$   ▷ add to mined set.
9:             **end if**
10:         **end for**
11:     **end for**
12:     $G_{t+1} = \texttt{train}\,(\mathcal{S} \cup \mathcal{B})$
13:     $L_{t+1}^{min} = L_t^{min} - 1$   ▷ reduce length threshold.
14:     $t = t + 1$   ▷ track iteration
15: **end while**

earlier iterations. The algorithm converges when accuracy@1 stops increasing on a development set. We note that our bootstrapping approach is applicable to any transliteration generation model.

To ensure that high quality name pairs are added to the mined set $\mathcal{B}$ during bootstrapping, we use the following discovery constraints.

### 4.2 Discovery Constraints

A word-transliteration pair $(x, y)$ is added to the set of mined pairs $\mathcal{B}$, only if all the following constraints are satisfied,

1. $y \in \mathcal{N}$. i.e., $y$ belongs in the dictionary.

2. $P(y \mid x) > \delta^{min}$. The model is sufficiently confident about the transliteration.

3. The ratio of lengths $\frac{|y|}{|x|}$ should be close to the average ratio estimated from $\mathcal{S}$ (Matthews, 2007). We encode this using the constraint $|\frac{|y|}{|x|} - r(\mathcal{S})| \leq \epsilon$, where $\epsilon$ is a tunable tolerance and $r(\mathcal{S})$ is the average ratio in $\mathcal{S}$.

4. $|y| > L_t^{min}$. We found that false positives were more likely to be short hypotheses in early iterations. As the model improves with each iteration, $L_t^{min}$ is lowered to allow more new pairs to be mined.

We note that our bootstrapping algorithm can be formulated as an instance of constraint driven learning (Chang et al., 2007, 2012).

## 5 Experimental Setup

Unless otherwise specified, we evaluate all generation models using the best model prediction $\hat{y}$ using acc@1 against the reference transliteration $y^*$.

**Training and Evaluation Dataset** We use the train and development sets from the Named Entities Workshop 2015 (Duan et al., 2015) (NEWS2015) for Hindi (hi), Kannada (kn), Bengali (bn), Tamil (ta) and Hebrew (he) as our train and evaluation set.[4] The size of the train set was ~12k, 10k, 14k, 10k and 10k respectively, and all evaluation sets were ~1k.

For the low resource experiments, we subsample 500 examples from each train set in the NEWS2015 dataset using five random seeds and report the averaged results. We also set aside a 1k name pairs from the corresponding NEWS2015 train set of each language as development data. The foreign script portion of the remaining train data is used as $\mathcal{V}_f$ in the bootstrapping algorithm.

**Model and Tuning Details** We implemented Seq2Seq(HMA) using PyTorch.[5] We used 50 dimensional character embeddings, and single layer GRU (Cho et al., 2014) encoder with 20 hidden states for all experiments. The Adam (Kingma and Ba, 2014) optimizer was used with default hyper-parameters, a learning rate of 0.001, a batch size of 1, and maximum of 20 iterations in all experiments. Beam search used a width of 10. For low-resource experiments, all bootstrapping parameters were tuned on the development data set aside above. $L_0^{min}$ is chosen from $\{10, 15, 20, 25\}$.

**Name Dictionary** We use a name dictionary of 1.05 million names constructed from the English Wikipedia (dump dated 05/20/2017) by taking the list of title tokens in Wikipedia sorted by frequency, and removing tokens which appears only once.

### 5.1 Comparisons

We compare with the following generation models:

**P&R (Pasternack and Roth, 2009)** A probabilistic transliteration generation approach that learns latent alignments between substrings in the source and the target words. The model is trained to score all possible segmentation and their alignments, using an EM-like algorithm.

---
[4]Test set was not available since shared task concluded.
[5]github.com/pytorch

**DirecTL+ (Jiampojamarn et al., 2009)** A HMM-like discriminative string transduction model that predicts the output transliteration using many-to-many alignments between the source word and target transliteration. Following Jiampojamarn et al. (2009), we use the m2m-aligner (Jiampojamarn et al., 2007) to generate the many-to-many alignments, and the public implementation of DirecTL+ to train models.[6]

**RPI-ISI (Lin et al., 2016)** A transliteration approach that uses a language-independent entity linking system (Wang et al., 2015) to jointly correct and re-rank the hypotheses produced by the generation model. We compare to both the unconstrained inference (U) approach and the entity linking constrained inference (+EL) approach.

**Seq2Seq w/ Att** A sequence to sequence generation model which uses soft attention as described in (Bahdanau et al., 2015). This model does not enforce monotonicity at inference time, and serves as direct comparison for Seq2Seq(HMA).

## 6 Experiments

This section aims to analyze: **(a)** how effective is Seq2Seq(HMA) for transliteration generation when provided all available supervision (§6.1)? and **(b)** how effective is the bootstrapping algorithm in the low-resource setting when only 500 examples are available (§6.2)?

### 6.1 Full Supervision Setting

We compare Seq2Seq(HMA) with previous approaches when provided all available supervision, to see how it fares under standard evaluation.

Results in the unconstrained inference (U) setting (Table 2 top 5 rows) shows Seq2Seq(HMA), denoted by "Ours", outperforms previous approaches on Hindi, Kannada, and Bengali, with almost 3-4% gains. Improvements over the Seq2Seq with Attention (Seq2Seq w/ Att) model demonstrate the benefit of imposing the monotonicity constraint in the generation model. On Tamil and Hebrew, Seq2Seq(HMA) is at par with the best approaches, with negligible gap (∼0.3) in scores. Overall, we see that Seq2Seq(HMA) can achieve better (and sometimes competitive) scores than state-of-the-art approaches in full supervision settings. When comparing approaches which use constrained inference (Table 2, rows 6 and 7), we see that using dictionary-constrained inference (as in Ours(DC)) is more effective than using a entity-linking model for re-ranking (RPI-ISI + EL).

---
[6]https://code.google.com/p/directl-p

| Lang. → Approach ↓ | hi | kn | bn | ta | he | Avg. |
|---|---|---|---|---|---|---|
| **Full Supervision Setting (5-10k examples)** | | | | | | |
| Seq2Seq w/ Att (U) | 35.5 | 33.4 | 46.1 | 17.2 | 20.3 | 30.5 |
| P&R (U) | 37.4 | 31.6 | 45.4 | **20.2** | 18.7 | 30.7 |
| DirecTL+ (U) | 38.9 | 34.7 | 48.4 | 19.9 | 16.8 | 31.7 |
| RPI-ISI (U) | 40.3 | 29.8 | 49.4 | **20.2** | 21.5 | 32.2 |
| Ours(U) | **42.8** | **38.9** | **52.4** | 20.5 | **23.4** | **35.6** |
| **Approaches Using Constrained Inference** | | | | | | |
| RPI-ISI + EL | 44.8 | 37.6 | 52.0 | **29.0** | **37.2** | 40.1 |
| Ours(DC) | **51.8** | **43.3** | **56.6** | 28.0 | 36.1 | **43.2** |
| **Low-Resource Setting (500 examples)** | | | | | | |
| Seq2Seq w/ Att (U) | 17.0 | 13.6 | 14.5 | 6.0 | 9.5 | 12.1 |
| P&R (U) | 21.1 | 16.6 | 34.2 | 9.4 | 13.0 | 18.9 |
| DirecTL+ (U) | 26.6 | 25.3 | 35.5 | 11.8 | 10.7 | 22.0 |
| Ours(U) | 29.1 | 27.7 | 37.7 | 11.5 | 16.2 | 24.4 |
| **Ours(U) + Boot.** | **40.1** | **35.1** | **50.3** | **17.8** | **22.8** | **33.2** |

Table 2: Comparing different approaches on the NEWS 2015 dataset using acc@1 as the evaluation metric. "Ours" denotes the Seq2Seq(HMA) model, with (.) denoting the inference strategy. Numbers for RPI-ISI are from Lin et al. (2016).

### 6.2 Low-Resource Setting

In Table 2 (rows under "Low-Resource Setting"), we evaluate different models in a low-resource setting when provided only 500 name pairs as supervision. Results are averaged over 5 different random sub-samples of 500 examples.

The results clearly demonstrate that all generation models suffer a drop in performance when provided limited training data. Note that models like Seq2Seq with Attention suffer a larger drop than those which enforce monotonicity, suggesting that incorporating monotonicity into the inference step in the low-resource setting is essential. After bootstrapping our weak generation model using Algorithm 1, the performance improves substantially (last row in Table 2). On almost all languages, the generation model improves by at least 6%, with performance for Hindi and Bengali improving by more than 10%. Bootstrapping results for the languages are within 2-4% of the best model trained with all available supervision.

To better analyze the progress of the transliteration model during bootstrapping, we plot the accuracy@1 of the current transliteration model after each bootstrapping iteration for each of the languages (solid lines in Figure 2). For reference, we also show the best performance for a gener-ation model using all available supervision from §6.1 (dotted horizontal lines in Figure 2). From Figure 2, we can see that almost after 5 bootstrapping iterations, the generation model attains competitive performance to respective state-of-the-art models trained with full supervision.

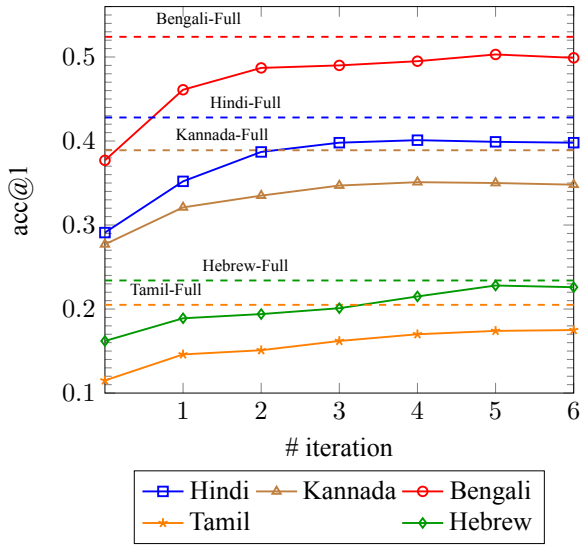

Figure 2: Plot showing acc@1 after each bootstrapping iteration for Hindi, Kannada, Bengali, Tamil and Hebrew, starting with only 500 training pairs as supervision. For comparison, the acc@1 of a model trained with all available supervision is also shown (respective dashed lines, marked X-Full).

## 6.3 Error Analysis

Though our model is state of the art, it does present a few weaknesses. We have found that the dictionary sometimes misleads the model during constrained inference. For example, the correct transliteration "vidyul" of the Hindi विद्युल, is not present in the dictionary, but another hypothesis "vidul" is. Another issue comes from the proportion of native (i.e., from the source language) and foreign (i.e., from English or other languages) names in the training data. It is usually not the case that the source and target scripts have the same transliteration rules. For example, य in Hindi might represent *ya* in English or Hindi names, but *ja* in German. Similarly, while अ should be *a* in Hindi names, it could be any of a few vowels in English. The NEWS2015 dataset does not report a native/foreign ratio, but by our estimation, it is about 70/30 for each language. This native and foreign names dichotomy are some of the inherent challenges in transliteration, that we discuss in detail in the next section.

## 7 Challenges Inherent to Transliteration

The fact that all models in Table 2 perform well or poorly on the same languages suggests that most of the observed performance variation is the result of factors *intrinsic* to the specific languages. Here we analyze some challenges that are inherent to the transliteration task, and explain why the performance ceiling is well under 100% for all languages, and lower for languages like Tamil and Hebrew than the others.

### 7.1 Source and Target-Specific Issues

**Source-Driven** Some transliteration errors are due to ambiguities in the source scripts. For instance, the Tamil script uses a single character to denote {*ta*, *da*, *tha*, *dha*}, a single character for {*ka*, *ga*, *kha*, *gha*}, etc., while the rest of the Indian scripts have unique characters for each of these. Thus, names like *Hartley* and *Hardley* are entirely indistinguishable in Tamil but are distinguishable in the other scripts. We illustrate this problem by transliterating back and forth between Tamil and Hindi. When transliterating Hindi→Tamil, the model achieves an accuracy of 31%, which drops to 15% when transliterating Tamil→Hindi, suggesting that the Tamil script is more ambiguous.

The Hebrew script also introduces error because it tends to omit vowels or write them ambiguously, leaving the model to guess between plausible choices. For example, the word מלך could be transliterated *melech* "king" just as easily as *malach* "he ruled." When Hebrew does write vowels, it reuses consonant letters, again ambiguously. For example, ה can be used to express *a* or *e*, so שמונה can be either *shmona* or *shmone* "eight masculine/feminine". The script also does not reliably distinguish *b* from *v* or *p* from *f*, among others.

All languages run into problems when they are faced with writing sounds that they do not natively distinguish. For example, Hindi does not make a distinction between *w* and *v*, so both *vest* and *west* are written as वेस्ट in its script.

These script-specific deficiencies explains why all models struggle on Tamil and Hebrew relative to the others. These issues cannot be completely resolved without memorizing individual source-target pairs and leveraging context.

**Target-Driven** Some errors arise from the challenges presented by target script (here Latin script for English). To handle English's notoriously convoluted orthography, a model has to infer silent let-ters, decide whether to use *f* or *ph* for /f/; use *k*, *c*, *ck*, *ch*, or *q* for /k/, and so on. The problem is made worse because English is not the only language that uses Latin script. For example, German names like *Schmidt* should be written with *sch* instead of *sh*, and for French names like *Margot* and *Margeau* (which are pronounced the same), we have to resort to memorization. The arbitrariness extends into borrowings from the source languages as well. For example, the Indian name *Bangalore* is written with a silent-*e*, and the name *Lakshadweep* contains *ee*, instead of the expected *i*.

|         | Native | Foreign | Ratio |
|---------|--------|---------|-------|
| Hindi   | 45.1   | 31.4    | 1.44  |
| Bengali | 63.1   | 20.1    | 3.14  |
| Kannada | 42.6   | 23.1    | 1.84  |
| Tamil   | 24.3   | 05.2    | 4.67  |

Table 3: Acc@1 for native and foreign words for four languages (§7.2). Ratio is native performance relative to foreign.

## 7.2 Disparity between Native and Foreign

All these issues come together to create a performance disparity between *native* names, which are well-integrated into the source language etymologically (Indian names like *Jasodhara* or *Ramanathan* for Hindi), and *foreign* names (French *Grenoble* or Japanese *Honshu* for Hindi), which are not. The above datasets include an unspecified mix of native and foreign names. This is a problem since any model must learn essentially separate transliteration schemes for each.

To quantify the effect of this, we annotate native and foreign names in the test split of the four Indian languages, and evaluate performance for both categories. Table 3 shows that our model performs significantly better on native names for all the languages. A possible reason for is that the source scripts were designed for writing native names (e.g., Tamil script lacks separate {*ta*, *da*, *tha*, *dha*} characters because the Tamil language does not distinguish these sounds). Furthermore, foreign names have a wide variety of origins with their own conventions as discussed in §7.1. The performance gap is proportionally greatest for Tamil, likely due to its script.

## 8 Case Studies

In this section, we evaluate the practical utility of our approach in low-resource settings and for downstream applications through two case studies.

We first show that obtaining an adequate seed list is possible with a few hours of manual annotation (§8.1) from a *single* human annotator. We then show the positive impact that our approach has on a downstream task, by evaluating its contribution to candidate generation for Tigrinya and Macedonian entity linking (§8.2).

| Language   | Monolingual Corpus | Vocabulary |
|------------|--------------------|------------|
| Punjabi    | Corpus ILCI-II♠    | 30k        |
| Armenian   | TED♣               | 50k        |
| Tigrinya   | Habit Project♦     | 225k       |
| Macedonian | TED♣               | 60k        |

♦=habit-project.eu/wiki/TigrinyaCorpus,
♠=tdil-dc.in,
♣=github.com/ajinkyakulkarni14/
TED-Multilingual-Parallel-Corpus

Table 4: Corpora used for obtaining foreign vocabulary $\mathcal{V}_f$ for bootstrapping in the case studies in §8.1 and §8.2.

### 8.1 Manual Annotation

The manual annotation exercises simulate a low-resource setting with only a single human annotator is available. We judge the usability of the annotations by training models on them and evaluating the models on test sets of 1000 names each, obtained from Wikipedia inter-language links. For bootstrapping experiments, we use the corpora shown in Table 4 to obtain foreign vocabulary $\mathcal{V}_f$.

**Languages Studied** We investigate performance on two languages: Armenian and Punjabi.

Spoken in Armenia and Turkey, Armenian is an Indo-European language with no close relatives. It has Eastern and Western dialects with different spelling conventions. Armenian Wikipedia is primarily written in the Eastern dialect, while our annotator was a native Western speaker.[7]

Punjabi is an Indic language from Northwest India and Pakistan that is closely related to Hindi. Our annotator grew up primarily speaking Hindi.

**Annotation Guidelines** Annotators were given two tasks. First, they were asked to write two names and their English transliterations for each letter in the source script: one beginning with the letter and another containing it elsewhere. (e.g. "**J**ulia" and "Ben**j**amin" for the letter "j" if the source were English). The is done to ensure good coverage over the alphabet. Next, annotators were shown a list of English words and were asked to provide plausible transliteration(s) into the target script. The list had a mix of recognizable foreign (e.g., *Clinton*, *Helsinki*) and native names (e.g., *Sarkessian*, *Yerevan* for Armenian).

---

[7]The annotator produced Western Armenian which was mechanically mapped to "Eastern" by swapping five Armenian character pairs: ŋ/ɯ, ɰ/p, p/ɥ, ǎ/ð, ɓ/ℊ

| Lang. → Approach ↓ | Punjabi | Armenian |
|---|---|---|
| Ours(U) | 33.4 | 49.9 |
| Ours(U) + Bootstrapping | 44.5 | 55.8 |
| Annotation Time (hours) | 5 | 4 |

Table 5: Acc@1 using human annotated seed set and bootstrapping the Seq2Seq(HMA) model. Both languages perform well relative to the other languages investigated so far. Both annotation sub-tasks took roughly the same time.

We collected about 600 and 500 annotated pairs respectively for Armenian and Punjabi. Table 5 shows that the performance of the models trained on the annotated data is comparable to that on the standard test corpora for other languages. This show that our approach is robust to human inconsistencies and regional spelling variations, and that obtaining an adequate seed list is possible with just a few hours of manual annotation.

### 8.2 Candidate Generation (CG)

Since transliteration is an intermediate step in many downstream multilingual information extraction tasks (Darwish, 2013; Kim et al., 2012; Jeong et al., 1999; Virga and Khudanpur, 2003; Chen et al., 2006), it is possibly to gauge its performance extrinsically by the impact it has on such tasks. We use the task of *candidate generation* (CG), which is a key step in cross-lingual entity linking.

The goal of cross-lingual entity linking (McNamee et al., 2011; Tsai and Roth, 2016; Upadhyay et al., 2018) is to ground spans of text written in any language to an entity in a knowledge base (KB). For instance, grounding **[Chicago]** in the following German sentence to `Chicago_(band)`.[8]

**[Chicago]** *wird in Woodstock aufzutreten.*

The role of CG in cross-lingual entity linking is to create a set of plausible entities given a string while ensuring the correct KB entity belongs to that set. For the above German sentence, it would provide a list of possible KB entities for the string *Chicago*: `Chicago_(band)`, `Chicago_(city)`, `Chicago_(font)`, etc., so that entity linking can select the band. Foreign scripts pose an additional challenge for CG because they must be transliterated before they are passed on to candidate generation. For instance, any mention of "Chicago" in Amharic must first be transliterated from ሺካጎ.

Most approaches for CG use Wikipedia interlanguage links to generate the lists of candidates (Tsai and Roth, 2016). While recent approaches such as Tsai and Roth (2018) have resorted to name translation for CG, they require over 10k examples for languages written in non-Latin scripts, which is prohibitive for low-resource languages with little Wikipedia presence.

**Candidate Generation with Transliteration**
We evaluate the extent to which our approach improves recall of a naive CG baseline that generates candidates by performing exact name match. For each span of text to be linked (or *query mention*), we first check if the naive name matching strategy finds any candidates in the KB. If none are found, the query mention is back-transliterated to English, and at most 20 candidates are generated using a inverted-index from English names to KB entities. The evaluation metric is recall@20, i.e., if the gold KB entity is in the top 20 candidates. We use Tigrinya and Macedonian as our test languages.

**Tigrinya** is a South Semitic language related to Amharic, written in the Ethiopic script, and spoken primarily in Eritrea and northern Ethiopia. The Tigrinya Wikipedia has <200 articles, so we use inter-language links (∼7.5k) from the Amharic Wikipedia instead to extract 1k name pairs for the seed set. We use the monolingual corpus in Table 4 for bootstrapping and evaluate on the *unsequestered set* provided under the NIST LoReHLT evaluation, containing 4,630 query mentions.

The Ethiopic script is an alphasyllabary, where each character is consonant-vowel pair. For example, the character መ is *mä*, ሚ with a tail is *mi*, and ሞ with a line is *mo*. With 26 consonants and 8 vowels, this leads to a set of >200 characters creating a sparsity problem since each character has its own Unicode code point. However, the code points are organized so that they can be automatically split[9] into unique consonant and vowel codes *without explicitly understanding the script*. We assign arbitrary ASCII codes to each consonant and vowel so that መ/*mä* becomes "D 1" and ሞ/*mo* becomes "D 6." This consonant-vowel splitting (CV-split) reduces the number of unique input characters to 55.

---
[8]Translation: Chicago will perform at Woodstock.

[9]Consonant = Unicode / 8; Vowel = Unicode % 8

| Approach | Recall@20 |
|---|---|
| **Tigrinya** | |
| Name match (baseline) | 31.4 |
| Ours | 35.6 |
| Ours (CV-split) | 41.3 |
| **Ours (CV-split) + Bootstrapping** | **46.2** |
| **Macedonian** | |
| Name match (baseline) | 33.6 |
| Ours | 72.2 |
| **Ours + Bootstrapping** | **76.8** |

Table 6: Comparing candidate recall@20 for different approaches on Tigrinya and Macedonian. CV-split refers to consonant-vowel splitting. Using our transliteration generation model with bootstrapping yields the highest recall, improving significantly over a name match baseline.

**Macedonian** is a South Slavic language closely related to the languages of the former Yugoslavia and written in a local variant of the Cyrillic alphabet similar to Serbian's. We use the Macedonian test set constructed by McNamee et al. (2011) containing 1956 query mentions. A seed set of 1k name pairs was obtained from the inter-language Wikipedia links for Macedonian, and the monolingual corpus from Table 4 is used for bootstrapping.

**Candidate Generation Results** Table 6 shows the results for the two languages. For Tigrinya, candidate generation with transliteration improves on the baseline by 4.2%. Splitting the characters (CV-split) gives another 5.7%, and adding bootstrapping gives 4.9% more. Our approach yields an overall 14.8% improvement in recall over the baseline, showing that we can effectively exploit the little available supervision by bootstrapping. Macedonian yields more dramatic results, where transliteration provides 38.6% improvement (more than double the baseline), with bootstrapping providing another 4.6%. The differences between Tigrinya and Macedonian is likely due both to their test sets, corpora and writing systems.

## 9 Conclusion and Future Work

We presented a new transliteration generation model, namely Seq2Seq(HMA), and a new bootstrapping algorithm that can iteratively improve a weak generation model using constrained discovery. The model presented here achieves state-of-the-art results on typical training set sizes, and more importantly, works well in a low-resource setting with the aid of the bootstrapping algorithm. The key benefit of the bootstrapping approach is that it can "recover" most of the performance lost in the low-resource setting when little supervision is available by training with a smaller seed set, an English name dictionary, and a list of unannotated words in the target script. Additionally, our bootstrapping algorithm admits any generation model, giving it wide applicability. Through case studies, we showed that collecting an adequate seed list is practical with a few hours of annotation. The benefit of incorporating our transliteration approach in a downstream task, namely candidate generation, was also demonstrated. Finally, we discussed some of the inherent challenges of learning transliteration and the deficits of existing training sets.

There are several interesting directions for future work. Performing model combination, either by developing hybrid transliteration models (Nicolai et al., 2015) or by ensembling (Finch et al., 2016), can further improve low resource transliteration. Jointly leveraging similarities between related languages, such as writing systems or phonetic properties (Kunchukuttan et al., 2018), also shows promise for low-resource settings. Our analysis suggests value in revisiting "transliteration in context" approaches (Goto et al., 2003; Hermjakob et al., 2008), especially for languages like Hebrew. We would also like to expand on the analyses provided in §7 which uncover challenges inherent to the transliteration task, particularly the impact of the native/foreign distinction in the train and test data, the difficulties posed by specific scripts or pairs of scripts, and how these impact both back- and forward-transliteration. Recent work from Merhav and Ash (2018) suggests many useful analyses that we would like to incorporate.

## Acknowledgments

The authors thank Mitch Marcus, Snigdha Chaturvedi, Stephen Mayhew, Nitish Gupta, Dan Deutsch, and the anonymous reviewers for their useful comments. We are grateful to the Armenian and Punjabi annotators for help with the case studies.

This work was supported under DARPA LORELEI by Contract HR0011-15-2-0025, Agreement HR0011-15-2-0023 with DARPA, and an NDSEG fellowship for the second author. Approved for Public Release, Distribution Unlimited. The views expressed are those of the authors and do not reflect the official policy or position of the Department of Defense or the U.S. Government.